\documentclass[final]{cvpr}
\usepackage{times}
\usepackage{epsfig}
\usepackage{graphicx}
\usepackage{amsmath}
\usepackage{amssymb}

\usepackage{multirow}
\usepackage{tabu}
\usepackage[ruled,vlined]{algorithm2e}
\usepackage{algpseudocode}

\usepackage[pagebackref=true,breaklinks=true,colorlinks,bookmarks=false]{hyperref}

\def\cvprPaperID{6161} 
\def\confYear{CVPR 2021}

\graphicspath{{./figure/}{./}}
\begin{document}

\title{Understanding the Robustness of Skeleton-based Action Recognition under Adversarial Attack}

\author{He Wang$^1$\thanks{https://youtu.be/DeMkN3efp9s}, Feixiang He$^1$, Zhexi Peng$^2$, Tianjia Shao$^2$\thanks{Corresponding author}, Yong-Liang Yang$^3$, Kun Zhou$^2$, David Hogg$^1$\\
$^1$University of Leeds, UK\ \ \ \ $^2$State Key Lab of CAD\&CG, Zhejiang University, China\\
\ \ \ \ $^3$University of Bath, UK\\
{\small\{h.e.wang, scfh, D.C.Hogg\}@leeds.ac.uk, \{zhexipeng, tjshao, kunzhou\}@zju.edu.cn, y.yang@cs.bath.ac.uk}\\

}

\maketitle

\begin{abstract}
Action recognition has been heavily employed in many applications such as autonomous vehicles, surveillance, etc, where its robustness is a primary concern. In this paper, we examine the robustness of state-of-the-art action recognizers against adversarial attack, which has been rarely investigated so far. To this end, we propose a new method to attack action recognizers which rely on the 3D skeletal motion. Our method involves an innovative perceptual loss which ensures the imperceptibility of the attack. Empirical studies demonstrate that our method is effective in both white-box and black-box scenarios. Its \textbf{generalizability} is evidenced on a variety of action recognizers and datasets. Its \textbf{versatility} is shown in different attacking strategies. Its \textbf{deceitfulness} is proven in extensive perceptual studies. Our method shows that adversarial attack on 3D skeletal motions, one type of time-series data, is significantly different from traditional adversarial attack problems. Its success raises serious concern on the robustness of action recognizers and provides insights on potential improvements.
\end{abstract}

\section{Introduction}

The research in adversarial attack has proven that deep learning is vulnerable to certain imperceptible perturbation on data, leading to security and safety concerns \cite{Szegedy:2014:SRA}; meanwhile, adversarial attack has been useful in improving the robustness of classifiers \cite{Liao_2018_CVPR}. Starting from object recognition, the list of target tasks for adversarial attack has been rapidly expanding, now including face recognition \cite{Sharif16AdvML}, point clouds \cite{Xiang:2019:PointAdv}, 3D meshes \cite{Xiao_2019_MeshAdv}, etc. While adversarial attack on static data (images, geometries, etc.) has been well explored, its effectiveness on time-series has only been attempted under a few settings such as videos \cite{9063523, DBLP:journals/corr/abs-1803-02536}. In this paper, we look into another type of time-series data: 3D skeletal motion, for action recognition tasks.

Skeletal motion has been widely used in action recognition \cite{Du:2015:HRN}. It can greatly improve the recognition accuracy by mitigating issues such as lighting, occlusion and posture ambiguity. In this paper, we show that 3D skeletal motions are vulnerable to adversarial attack but their vulnerability is different from other data. The adversarial attack on 3D skeletal motion faces two unique and related challenges: low redundancy and perceptual sensitivity. When attacking images/videos, it is possible to perturb some pixels without causing too much visual distortion. This largely depends on the redundancy in the image space \cite{Tramr2017TheSO}.  Unlike images, which have thousands of Degrees of Freedom (DoFs), a skeletal motion is usually parameterized by fewer than 100 DoFs, i.e. the joints of the skeleton. This not only restricts the space of possible attacks \cite{Tramr2017TheSO}, but also affects the imperceptibility of the adversarial samples: a small perturbation on a single joint can be easily noticed. Furthermore, coordinated perturbations on multiple joints in only one frame can hardly work either, because in the temporal domain, similar constraints apply. Any sparsity-based perturbation (on single joints or individual frames) will greatly affect the dynamics (causing jittering or bone-length violations) and will be very obvious to an observer. One consequence is that the perturbation magnitude alone is not anymore a reliable metric to judge the imperceptibility of an attack, as an overall small perturbation could still break the dynamics. This is very different from existing attack tasks where the perturbation magnitude can be heavily relied upon.

To systematically investigate the robustness of action recognizers, we propose a straightforward yet very effective method, Skeletal Motion Action Recognition Attack (SMART), based on an optimization framework that explicitly considers motion dynamics and skeletal structures. The optimization finds perturbations by balancing between classification goals and perceptual distortions, formulated as classification loss and perceptual loss. Varying the classification loss leads to different attacking strategies. The new perceptual loss fully utilizes the dynamics of the motions and bone structures. SMART is effective in both white-box and black-box settings, on several state-of-the-art models, across a variety of datasets.

Formally, we systematically investigate the vulnerability of a wide range of state-of-the-art methods under adversarial attack and identify their weaknesses for potential improvements. To this end, we propose a new adversarial attack method with a novel perceptual loss function capturing the perceptual realism and fully exploiting the motion dynamics. We also provide insights into the role of dynamics in the imperceptibility of the adversarial attack based on comprehensive perceptual studies, showing that it is not enough to only constrain the perturbation magnitude, which differs significantly from widely accepted approaches.

\section{Related Work}
\subsection{Skeleton-based Action Recognition}
Action recognition is crucial in many applications, namely surveillance, human-robot interaction and entertainment. Recent advances in 3D sensing and pose estimation motivate the use of clean skeleton data to robustly classify human actions, overcoming the biases in raw RGB videos due to body occlusion, scattered background, lighting variation, etc. Unlike conventional approaches that are limited to handcrafted skeletal features~\cite{Vemulapalli:2014:HAR,Fernando:2015:MVE,Devanne:2015:HAR}, recent methods taking the advantage of trained features from deep learning have gained state-of-the-art performance. Based on the representation of skeletal data, deep learning based methods can be classified into three categories, including sequence-based, image-based, and graph-based methods.

Sequence-based methods represent a skeletal motion as a chronological sequence of poses, each of which consists of the coordinates of all the joints. Then RNN-based architecture is employed to perform the classification~\cite{Du:2015:HRN,Liu:2016:STL,Song:2017:ESA,Zhang:2019:VAN}. Image-based methods represent a skeletal motion as a pseudo-image, which is a 2D tensor where one dimension corresponds to time, and the other dimension stacks all the joints of a single skeleton. Such representation enables CNN-based image classification to be applied to action recognition~\cite{Liu:2017:ESV,Ke:2017:NRS}. Different from the previous two categories that mainly rely on skeleton geometry represented by the joint coordinates, graph-based methods utilize graph representations to naturally consider the skeleton topology (i.e. joint connectivity) which is encoded by bones that connect neighboring joints. Graph neural networks (GNN) are then used to recognize the actions~\cite{Shi:2019:SBA,Cheng_2020_CVPR,Liu_2020_CVPR,Zhang_2020_CVPR_1,Zhang_2020_CVPR_2}. Based on the code released by the authors, we perform adversarial attacks on the two most representative categories (i.e. RNN- and GNN-based), demonstrating the vulnerability of existing methods.

\subsection{Adversarial Attacks}
Despite their significant successes, deep neural networks are vulnerable to carefully crafted adversarial attacks as firstly identified in~\cite{Szegedy:2014:SRA}. Delicately designed neural networks with high performance can be easily fooled by unnoticeable perturbations on the input data. With the concern raised, researchers have extensively investigated adversarial attacks on different data types, including 2D images~\cite{Goodfellow:2014:explaining,Papernot:2015:LDL,Dezfooli:2016:DeepFool,Xiao:2018:GAE,Xiao:2018:Spatially}, videos~\cite{Wei:2018:SparseAP,Wang:2018:LDV}, 3D shapes~\cite{Liu:2019:BPN,Zeng:2019:AAB,Xiao_2019_MeshAdv,Xiang:2019:PointAdv}, physical objects~\cite{Kurakin:2016:AEP,Athalye:2017:SRA,Evtimov:2017:RPW}, graphs~\cite{pmlr-v80-dai18b}, while little attention has been paid to 3D skeletal motions.

The adversarial attack in the context of action recognition is much less explored. Inkawhich et al.~\cite{Inkawhich:2018:AAO} perform adversarial attacks on optical-flow based action classifiers, which is mainly inspired by image-based attacks and differs from our work in terms of the input data. The adversarial attack on skeletal motions has just been attempted recently \cite{Liu:2019:AAS,Zheng2020TowardsUT} (arXiv only). However, they did not investigate the imperceptibility systematically, which is crucial as shown in our perceptual studies because imperceptibility is a strong requirement on adversarial attack. In our work, we demonstrate better results using a perceptual loss that minimizes the motion derivative deviation relative to the original skeletal motion, thereby preserving the motion dynamics which are intrinsic to actions. This is crucial in attacking highly dynamic motions such as running and jumping. We also perform a perceptual study to systematically validate the imperceptibility of the perturbed skeletal motions and the effectiveness of our choice of perceptual loss.

We demonstrate successful attacks on a range of network architectures, including RNN and GNN based methods, on three datasets. Finally, we present results of three different attacking strategies, including the novel objective of placing the correct action beneath the first n actions in a ranked classification, for a given n.


\section{Methodology}
SMART is formulated as an optimization problem, where the minimizer is an adversarial sample, for a given motion, that minimizes the perceptual distortion while fooling the target classifier. The optimization has variants constructed for three different attacking strategies: \textit{Anything-but Attack}, \textit{Anything-but-N Attack} and \textit{Specified Attack}. They are used in \textit{white-box} and \textit{black-box} scenarios.

\subsection{Optimization for Attack}
Given a motion $q$ = \{$q_0$, $q_1$, ..., $q_t$\}, where $q_t$ is the frame at time $t$ and consists of stacked 3D joint locations, a trained classifier $\Phi$ can predict its class label $y_q$ = $C(\Phi(q)$), where $\Phi$ is namely a deep neural network and $\Phi(q)$ is the predicted distribution over class labels. $C$ is usually a \textit{softmax} function and $y_q$ is the predicted label. We aim to find a perturbed example, $\hat{q}$, for $q$, such as $y_q \neq y_{\hat{q}}$. A common method is to find \textit{the minimal perturbation} \cite{Xu2020AdversarialAA} through solving a constrained optimization. We start with the C\&W formulation\cite{7958570}:
\begin{align}
\label{eq:cw}
    \text{min}\ L_p(q, \hat{q}) \ 
    \text{sub. to}\ C(\Phi(\hat{q})) = c\  \text{and}\ \hat{q} \in [0,1]^n
\end{align}
where $L_p$ is a distance function and $C$ is a hard constraint dictating that the predicted class of $\hat{q}$ (bounded in $[0,1]^n$) being $c$. However, directly solving Eq. \ref{eq:cw} is difficult due to that $C$ is highly non-linear \cite{7958570}. So it can be relaxed by moving the hard constraint into the objective:
\begin{equation}
\label{eq:relaxed}
    \text{minimize}\ L = w L_c(y_{\hat{q}}, c) + (1-w)L_p(q, \hat{q})
\end{equation}
where $L_c$ is a classification loss and $w=0.4$. $L_p$ is normally the perturbation magnitude \cite{7958570}. But we use a new perceptual loss which is explained later. Eq.\ref{eq:relaxed} has intuitive interpretation: there are two forces governing $\hat{q}$. $L_c$ is the classification loss (a relaxed $C$ in Eq.\ref{eq:cw}) where we can design different attacking strategies. $L_p$ is the perceptual loss which dictates that $\hat{q}$ should be visually indistinguishable from $q$. To optimize for $\hat{q}$, we have only one assumption: we can compute the gradient: $\frac{\partial L}{\partial \hat{q}}$. This way, we can compute $\hat{q}$ iteratively by $\hat{q}_{t+1}$ = $\hat{q}_{t}$ + $\epsilon f(\frac{\partial L}{\partial \hat{q}_t}, \hat{q}_t)$ where $\hat{q}_t$ is $\hat{q}$ at step $t$, $f$ computes the updates and $\epsilon$ is the learning rate. We set $\hat{q}_0$ = $q$ and use Adam \cite{Diederik_2014} for $f$.

\subsection{Perceptual Loss}
\label{sec_pl}
Imperceptibility (governed by $L_p$ in Eq.\ref{eq:relaxed}) is a hard constraint in adversarial attacks. It requires that human cannot distinguish easily between the adversarial samples and real data. Existing approaches on images and videos achieve imperceptibility by constraining the pixel-wise or frame-wise perturbation magnitude measured by $l$ norms. One major difference in our problem is motion dynamics.

To fully represent the dynamics of a motion, we need the derivatives from \textit{zero-order} (joint location), \textit{first-order} (joint velocity)  up to \textit{nth-order}. One common approximation is to use first $n$ terms. When it comes to imperceptibility, the perceived motion naturalness is vital and not all derivatives are at the same level of importance \cite{Wang_STRNN_2019}. Inspired by the work in character animation \cite{wang_energy_2015,wang_harmonic_2013,Chen_dynamic_20}, we propose a new perceptual loss:
\begin{eqnarray}
    L_p(q, \hat{q}) = \alpha l_{dyn} + (1-\alpha)l_{bl} \\
    l_{bl} = ||Bl(q) - Bl(\hat{q})||_2^2 = \frac{1}{M}\sum_{i=1}^{M} ||Bl(q_i) - Bl(\hat{q}_i)||_2^2\\
    l_{dyn} = \sum_{n=0}^{\infty} \beta_n||(q^n - \hat{q}^n)||_2^2~~\textrm{where} \sum_{n=0}^{\infty} \beta_n = 1
\end{eqnarray}
where $\alpha$ = 0.3. $l_{bl}$ penalizes any bone length deviations in every frame where $M$ is the total frame number. $Bl(q_i) \in$ $\mathbb{R}^{24\times1}$ is the bone length vector of frame $q_i$. Theoretically, bone lengths do not change over time. However, they do vary in the original data due to tracking errors. This is why $l_{bl}$ is designed to be frame-wise.

$l_{dyn}$ is the dynamics loss. We use a strategy called \textit{derivative matching}. It is a weighted (by $\beta_n$) sum of the $l_2$ distance between $q^n$ and $\hat{q}^n$, where $q^n$ and $\hat{q}^n$ are the $nth$-order derivatives and can be computed by forward differencing. Although $n$ goes up to infinity, in practice, we explored up to $n=4$, which includes joint position, velocity, acceleration, jerk and snap. After exhaustive experiments, we find that enforcing the $0th$, $2nd$ and $4th$ order derivatives while discarding other derivatives gives good results, with the $4th$ derivative adding small gains. Including consecutive derivatives (e.g. $0th$, $1st$ and $2nd$) over-constrains the system. Also, the gain of including higher order derivatives diminishes while incurring more computation. A good compromise is to set $\beta_0 = 0.6$ and $\beta_2 = 0.4$. Matching the 2$nd$-order profiles of two motions is critical. For skeletal motions, small location deviations can still generate large acceleration differences, resulting in two distinctive motions. More often, it generates severe jittering and thus totally unnatural motions. An alternative way of regulating the dynamics is to purely smooth the motion, by e.g. minimizing the acceleration. But it dampens highly dynamic motions such as jumping~\cite{Wang_STRNN_2019}. Also, considering more derivatives above $n=4$ makes the optimization harder to solve and over-weighs their benefits.


\subsection{White-box Attack}
\label{sec:whitebox}
With the perceptual loss designed, varying the formulation of the classification loss ($L_c$ in Eq.\ref{eq:relaxed}) allows us to form different attacking strategies. We present three strategies.

{\bf Anything-but Attack (AB)} aims to fool the classifier so that $y_q \neq y_{\hat{q}}$. This can be achieved by maximizing the \textit{cross entropy} between $\Phi(q)$ and $\Phi(\hat{q})$:
\begin{equation}
    L_c(q, \hat{q}) = -cross\_entropy(\Phi(q), \Phi(\hat{q}))
\end{equation}

{\bf Anything-but-N Attack (ABN)} is a generalization of AB. It aims to confuse the classifier so that it has similar confidence levels in multiple classes. ABN is more suitable to confuse classifiers which rely on top N accuracy. In addition, we find that it performs better in black-box attacks by transferability, which will be detailed in experiments. One naive solution is to use multiple AB losses for the top $n$ classes, but it will make the optimization difficult and will not scale as the class number increases. Instead, we propose an easier loss function, maximizing the entropy of the predicted distribution of $\hat{q}$:
\begin{eqnarray}
    L_c(q, \hat{q}) = -Entropy(\Phi(\hat{q})), \ \ \ \ 
    y_q \not\in TopN(\Phi(\hat{q}))
\end{eqnarray}
where $TopN$ is the set of the top $n$ class labels in the predictive distribution $\Phi(\hat{q})$. By minimizing $L_c$, we actually maximize the entropy of $\Phi(\hat{q})$, i.e. forcing it to be flat over all the class labels and thus reduce the confidence of the classifier over any particular class. We stop the optimization once the ground-truth label falls beyond the top $n$ classes. ABN is a harder optimization problem than AB because it needs the predictive distribution to be as flat as possible.

\subsubsection{Specified Attack (SA)}
Different from AB and ABN, sometimes it is useful to fool the classifier with a pre-defined class label. Given a fake label $y_{\hat{q}}$, we can use its class label distribution $\Phi_{\hat{q}}$, a one-hot vector, and minimize the cross entropy:
\begin{equation}
    L_c(q, \hat{q}) = cross\_entropy(\Phi(q), \Phi(\hat{q}))
\end{equation}
This is the most difficult scenario because it highly depends on the similarity between the source and target label. While turning `clapping over the head' into `raising two hands' is achievable with minimal visual changes, turning `running' into `squat' without being noticed is much harder.

\subsection{Black-box Attack}
While the white-box attack relies on the ability to estimate $\frac{\partial L}{\partial \hat{q}}$, which requires the access to the target classifier and is not always possible, black-box attack assumes that the full knowledge of the target classifier is inaccessible. We therefore cannot directly compute $\frac{\partial L}{\partial \hat{q}}$. Under such circumstances, we use \textit{attack-via-transferability} \cite{Tramr2017TheSO}. It begins with training a surrogate classifier. Then adversarial samples are computed by white-box attacks on the surrogate classifier. Finally, the adversarial samples are used to attack the target classifier in a black-box setting. In this paper, we do not construct our own surrogate model. Instead, we use an existing classifier as our surrogate classifier to attack others. In experiments, we attack several state-of-the-art models. To test the transferability and generalizability of our method, we use every model in turn as the surrogate model and attack the others.

\section{Experimental Results}

We first introduce the datasets and models for our experiments, followed by our white-box and black-box results. We then present our perceptual studies on the imperceptibility and compare SMART with other methods. During the attack, we first use the source code shared by the authors if available or implement the methods ourselves. Then we train them strictly following the protocols in their papers. Next, we test the models and collect the data samples that the trained classifiers can successfully recognize, to create our adversarial attack datasets. Finally, we compute the adversarial samples using different attacking strategies.

\subsection{Datasets}
\label{exp_data}
We choose three widely used datasets. \textbf{HDM05} ~\cite{cg-2007-2} contains 2337 sequences for 130 actions performed by 5 non-professional actors. The 3D joint locations of the subjects are provided in each frame. \textbf{MHAD}~\cite{6474999} is captured using a multi-modal acquisition system, consisting of 11 actions performed by 12 subjects, where 5 repetitions are performed for each action, resulting in 659 sequences. In each frame, the 3D joint positions are extracted based on the 3D marker trajectories. \textbf{NTU60}~\cite{Shahroudy_2016_NTURGBD} is captured by Kinect v2 and is currently one of the largest publicly available datasets for 3D action recognition. It is composed of more than 56,000 action sequences. A total of 60 action classes are performed by 40 subjects. The 3D coordinates of joints are provided by Kinect. Due to the huge number of samples and the large intra-class and viewpoint variations, the NTU60 is very challenging and is highly suitable to validate the effectiveness and generalizability of our approach. Note we exclude Kinectics \cite{Kay_Kinetics_2017}, a dataset that is also used in many papers, for two reasons. First, some older recognizers we investigate cannot achieve reasonable classification accuracy on it. Second, its quality is too low to evaluate the success of the attack, explained in Section \ref{exp_study}.

\subsection{Target Models}
\label{exp_model}
Rather than focusing only on the most recent methods, we select a range of methods: HRNN~\cite{Du_CVPR_2015}, ST-GCN~\cite{Yan_AAAI_2018}, AS-GCN~\cite{Li_2019_CVPR}, DGNN~\cite{Shi:2019:SBA}, 2s-AGCN~\cite{Shi_2019_CVPR}, MSG3D~\cite{Liu_2020_CVPR_1} and SGN~\cite{Zhang_2020_CVPR_3}, and investigate their vulnerability under different scenarios. They include both RNN- and GNN-based models. We implement HRNN following the paper and use the code shared online for the rest of the methods. We also follow their protocols in data pre-processing. Specifically, we preprocess the HDM05 and MHAD as in~\cite{Du_CVPR_2015} (where HDM05 is grouped into 60 classes), and the NTU60 as in~\cite{Shi_2019_CVPR}. We also map different skeletons to a standard 25-joint skeleton as in \cite{Wang_STRNN_2019}. 




\subsection{White-box Attack}
\label{exp_white}
In this section, we qualitatively and quantitatively evaluate the performance of SMART. We use a learning rate between 0.005 and 0.0005 and a maximum of 300 iterations. The setting for AB and ABN is straightforward. In SA, the number of experiments needed would be prohibitively large if we were to attack every motion with every other label but the ground-truth. Instead, we randomly select fake labels to attack. Since the number of motions attacked is large, the results are sufficiently representative. Note that this is a very strict test as most of the motions are rather distinctive. For simplicity, we only show representative results in the paper. For more results, please refer to the supplementary materials and video.



\subsubsection{Attack Results.}


We show the quantitative results of AB in Table \ref{table_attack} Left. High success rates are universally achieved across different datasets and target models, demonstrating the generalizability of SMART. For adversarial attack, it is not surprising if the before-attack and after-attack labels are semantically similar, e.g. from drinking water to eating. In SMART, a variety of examples are found where the after-attack labels are significantly different from the original ones. Due to the space limit, we leave all the details in the supplementary video and materials and only give a couple of examples here.  In HDM05, high confusion is found between turn\_L (turn left) and walk\_rightRC (walk sideways, to the right, feet cross over alternately front/back) in HRNN. Similarly, in NTU, high confusion is found between standing\_up (from sitting) and wear\_a\_shoe in 2SAGCN. These labels have completely different semantics and involve different body parts and motion patterns. Moreover, this kind of confusion is observed across all datasets and models. 


\begin{table*}[tb]
	\centering

		\begin{tabular}{c|ccc|ccc|ccc}		
			\hline
			Model/Data & HDM05 & MHAD  & NTU & HDM05 & MHAD  & NTU & HDM05 & MHAD  & NTU \\ \hline
			HRNN  & 100   & 100 & 99.56 & 100/100 & 100/100 & 99.84/99.62 & 67.19  & 57.41 & 49.17      \\
			ST-GCN & 99.57   & 99.96 & 100 & 93.30/90.28 & 76.86/70.5 & 95.86/91.32 & 74.95 & 66.93 & 100      \\
			AS-GCN & 99.36   & 92.84 & 97.43 & 91.46/82.83 & 42.07/22.34 & 91.18/82.47 & 64.62 & 40.18 & 99.48    \\
			DGNN & 96.09   & 94.46 & 92.51 & 93.55/86.32 & 87.54/74.27 & 98.73/97.62 & 97.26 & 96.13 & 99.99 \\
			2s-AGCN & 99.18   & 95.97 & 100 & 83.40/75.2 & 55.9/32.08 & 100/100 &  96.72 & 97.53 & 100  \\
			\hline
			mean & 98.84 & 96.65 & 97.9 & 92.34/86.93 & 72.47/59.84 & 97.12/94.21 & 80.15 & 71.64 & 89.73     \\ \hline
		\end{tabular}
	\caption{Success rate. Left: Anything-but (AB) Attack. Mid: Anything-but-N Attack. The results are AB3/AB5 when n = 3 (AB3) and 5 (AB5). Right: Specified Attack (SA).}
	\label{table_attack}
\end{table*}

We show the ABN results in Table~\ref{table_attack} Mid, in two variations: AB3 and AB5, as a generalization of AB. They are good for attacking classifiers based on top N accuracy. ABN is a harder problem than AB, with AB5 being harder than AB3, hence has a lower success rate. In terms of datasets, MHAD is the hardest for ABN because there are only 11 classes as opposed to 65 and 60 in the other two. Excluding the ground-truth label from the top 5 out of 11 classes is much more challenging than that of 65 and 60 classes.


Table \ref{table_attack} Right shows the SA results. SA is the most difficult because randomly selected class labels often come from significantly different action classes. Although it might be easy to confuse the model between `deposit' and `grab', it is extremely difficult to do so for `jumping' and `wear-a-shoe'. However, even under such circumstances, SMART is still able to succeed in more than 70\% cases on average, with multiple tests above 96\% and even achieving 100\%.

\textbf{Performance.} The major computational cost comes from the gradient estimation which depends on the target model because it requires back-propagation. We run a maximum of 300 iterations. The total amount of time each iteration takes are on average 0.102s, 0.267s, 0.419s, 0.275s and 0.738s on HRNN, ST-GCN, AS-GCN, DGNN and 2S-AGCN respectively, on Nvidia GTX 1080Ti (DGNN and 2S-AGCN) and TitanXp (HRNN, ST-GCN and AS-GCN).



\subsection{Black-box Attack}
\label{exp_black}
In the black-box setting, we attack the NTU dataset. Since we need a surrogate model to fool the target models, we first use 2s-AGCN as the surrogate model to attack DGNN, AS-GCN, MSG3D and SGN. The results are shown in Table \ref{table_bb}. We notice that SMART achieves successes on all target models except MSG3D, which indicates that not all target models are equally easy to fool by the transferred black-box attack. To further investigate it, we use three models: AS-GCN, DGNN and 2s-AGCN, and in turn take every model as the surrogate model and produce adversarial examples using AB and AB5. 

\begin{table}[tb]
	\centering
	\begin{tabular}{ccccc}		
		\hline
		DGNN  & AS-GCN & MSG3D & SGN              \\
		\hline
		98.37  & 98.10 & 3.08\% & 97.75\%  \\
		\hline
	\end{tabular}

	\caption{Success rate of AB black-box attack, using 2s-AGCN.}
	\label{table_bb}
\end{table}

\begin{table}[tb]
	\centering
	\begin{tabular}{c|ccc}		
		\hline
		 & DGNN & 2s-AGCN  & AS-GCN              \\ \hline
		DGNN                    & n/a  & 90.6(90.99) & 7.24(7.63)        \\
		2s-AGCN                      & 98.37(98.46)   & n/a & 98.10(98.96)      \\
		AS-GCN                    & 10.90(12.97)   & 91.17(91.99) & n/a      \\
		\hline
	\end{tabular}

	\caption{Success rate (AB/AB5) of black-box attack.}
	\label{table_mbb}
\end{table}

\begin{figure*}[tb]
        \centering
        \includegraphics[width=0.8\linewidth]{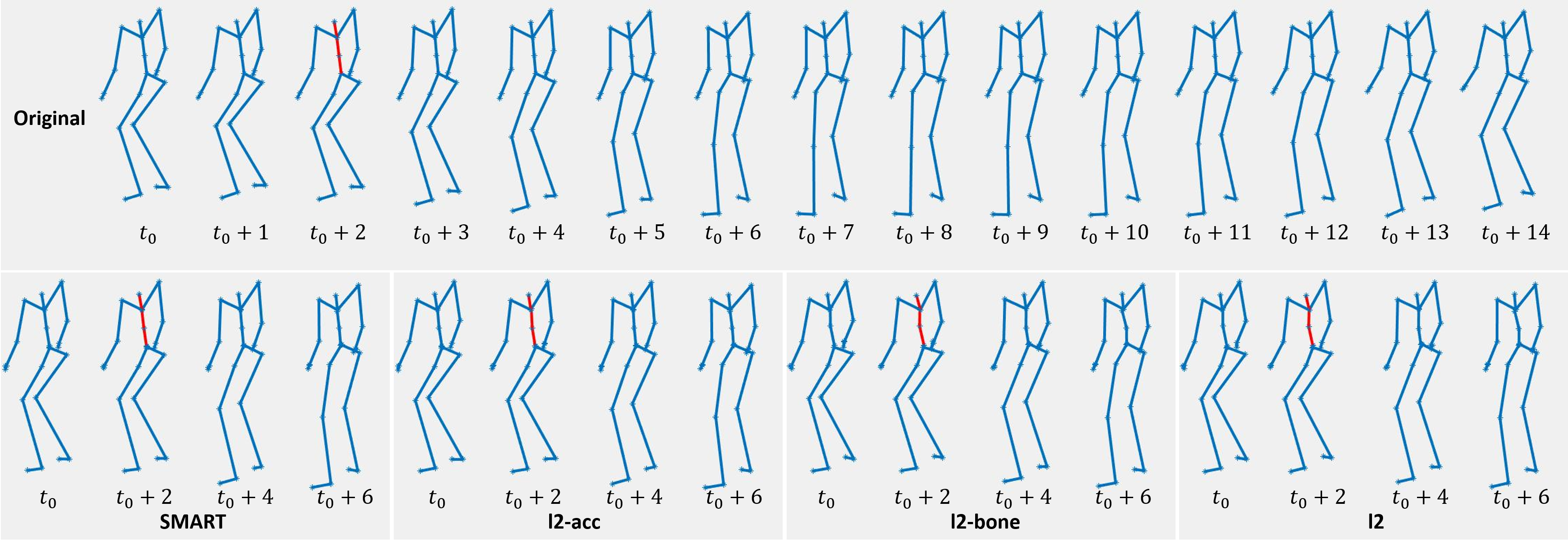}


  \caption{Visual comparison between different losses. Highlighted spine areas in the same frame show key visual differences.}
  \label{fig:ablation}
\end{figure*}

Results are shown in Table \ref{table_mbb}. AB5 results are in general better than AB. We speculate that there are two factors.  First, the predictive class distribution of AB5 is likely to be flatter than AB. The flatness improves the transferability because a target model with similar decision boundaries will also produce a similarly flat predictive distribution, and thus is more likely to be fooled. Besides, since the ground-truth label is pushed away from the top 5 classes in the surrogate model, it is also likely to be far away from the top in the target model. We also notice that the transferability is not universally successful. DGNN and AS-GCN cannot easily fool one another. Meanwhile, 2S-AGCN can fool and be fooled by both of them. Since the transferability can be described by distances between decision boundaries \cite{Tramr2017TheSO}, our speculation is that 2S-AGCN's boundary structure overlaps with both DGNN and AS-GCN significantly but the other two overlap little. The theoretical reason is hard to identify, as the formal analysis on transferability has just emerged on static data \cite{Tramr2017TheSO,Zhao_AAAI_2019}. The theoretical analysis of time-series data is beyond the scope of this paper and is therefore left for future work.

\subsection{Perceptual Study}
\label{exp_study}

One key difference between SMART and existing work is that we employ both \textit{numerical accuracy} and \textit{rigorous perceptual studies} to evaluate the success of attacks. Imperceptibility is a requirement for any adversarial attack. All the success shown above would have been meaningless if the attack were noticeable to humans. To evaluate imperceptibility, qualitative visual comparisons can be used on the image-based attack, but rigorous perceptual studies are needed for complex data \cite{Xiao_2019_MeshAdv}, as the numerical success can always be achieved by sacrificing the imperceptibility. This is especially the case for motions. Also, the necessity of perceptual studies restricts us from using noisy datasets (e.g. Kinetics \cite{Kay_Kinetics_2017}) because the subjects are unable to identify perturbations in side-by-side comparisons due to the excessive jittering and tracking errors in the original data.

We conduct three user studies (Deceitfulness, Naturalness and Indistinguishability). Since our sample space is huge (7 models $\times$ 3 datasets $\times$ 3 attacking strategies), we choose the most representative setting. We use the adversarial samples under AB in HDM05 and MHAD. NTU dataset is only used in visual evaluation, not perceptual study due to motion jittering in the original data (see the video for details). In total, we recruited totally 41 subjects (age between 18 and 37). Details are in the supplementary materials.

{\bf Deceitfulness}. In each user study, we randomly choose 100 motions with the ground-truth and the after-attack label for 100 trials. In each trial, the video is played for 6 seconds and then the subject is asked to choose which label best describes the motion with no time limit. This is to test whether SMART visually changes the semantics of the motion. This is also to test whether people can distinguish actions by only observing skeletal motions.

{\bf Naturalness}. Since unnatural motions can be easily identified as a result of the attack, we perform ablation tests on different loss term combinations. We design four settings: l2, l2-acc, l2-bone, SMART. l2 is where only the $l_2$ norm of joint perturbation is used, which is also widely used in existing methods such as image/video/mesh attack. l2-acc is l2 plus the acceleration loss, l2-bone is l2 plus the bone-length loss and SMART is the proposed perceptual loss. We first show static poses in Figure \ref{fig:ablation}. Motion comparisons are available in the supplementary video. Visually, SMART is the best. Even from static poses, one can easily see the artifacts caused by joint displacements. The spinal joints are the most obvious. The joint displacements cause unnatural zig-zag bending in l2, l2-acc and l2-bone, which is even more obvious in motions.

Next, we conduct perceptual studies. In each study, we randomly select 50 motions. For each motion, we make two trials. The first includes one attacked motion by SMART and one randomly selected from l2, l2-acc and l2-bone. The second includes two motions randomly drawn from l2, l2-acc and l2-bone. The first trial evaluates our results against other alternatives and the second reveals the impact of different perceptual loss terms. In each of the 100 trials, two motions are played together for 6 seconds twice, and then the subject is asked to choose which motion looks more natural or cannot tell the difference, with no time limit.

{\bf Indistinguishability}. In this study, we conduct a very strict test to see if the users can tell if a motion is perturbed in any way at all. In each experiment, 100 pairs of motions are randomly selected. In each trial, the left motion is always the original and the user is told so. The right one can be the original ({\bf sensitivity}) or attacked ({\bf perceivability}). We ask if the user can see any visual differences. Each video is played for 6 seconds then the user is asked to choose if the right motion is a changed version of the left, with no time limit. This user study serves two purposes. Perceivability is a direct test on Indistinguishability on the attack while sensitivity is to screen out subjects who tend to give random choices. Most users are able to recognize if two motions are the same (close to 100\% accuracy), but there are a few whose choices are more random. We discard any user data which falls below 80\% accuracy on the sensitivity test.

\begin{figure*}[tb]
  \centering
  \includegraphics[width=0.4\linewidth]{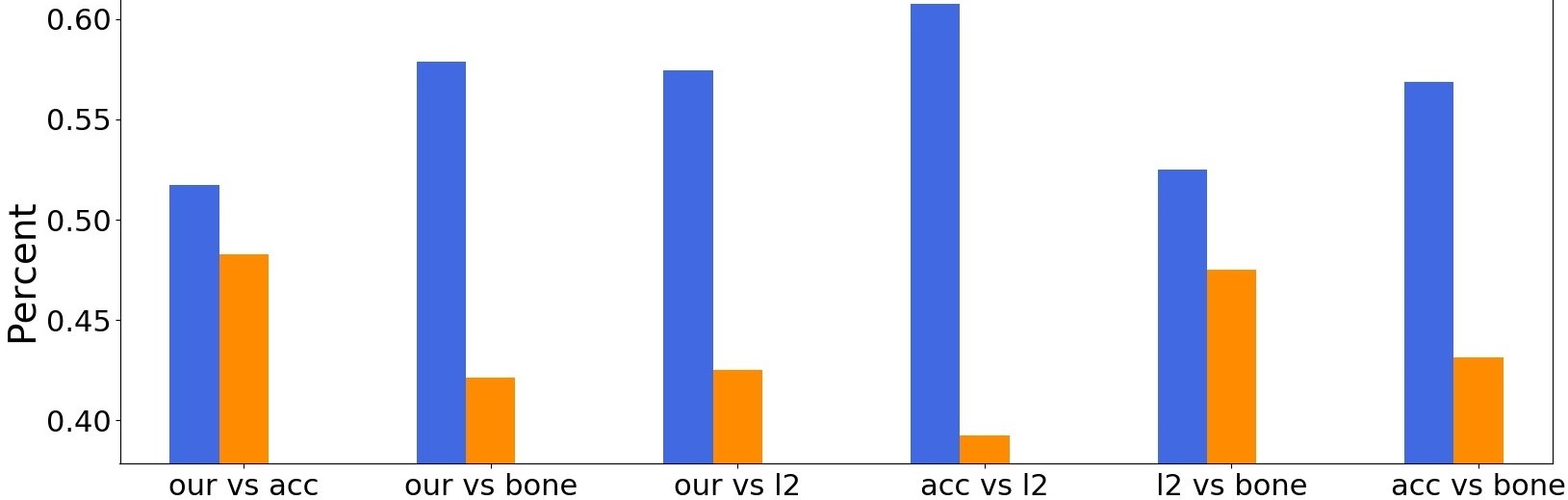} \ \ \
  \includegraphics[width=0.4\linewidth]{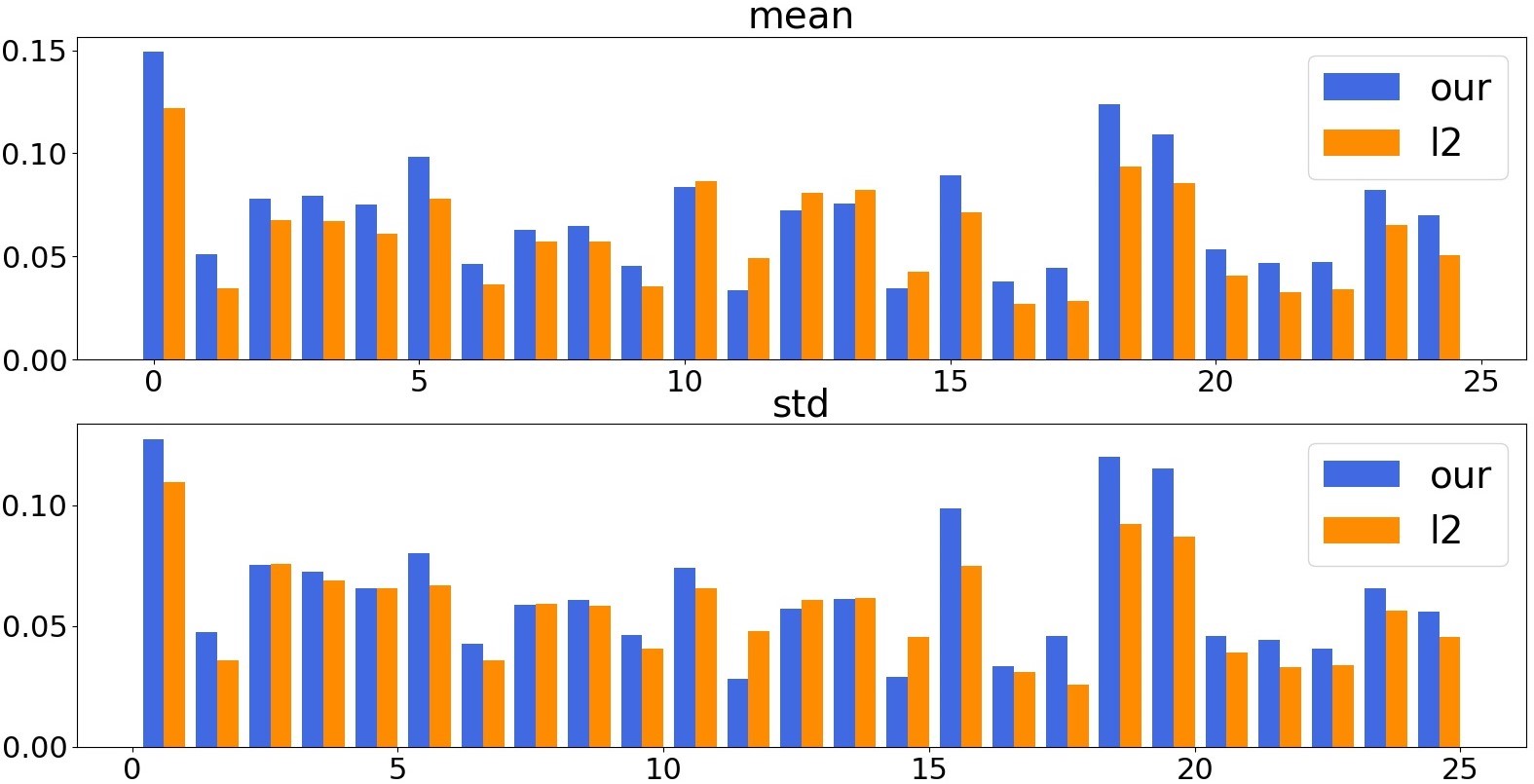}
  \caption{Left: Normalized user preference on Naturalness. our: SMART. bone: l2-bone. acc: l2-acc. The vertical axis is the percentage of user preference. Right: The mean (Top) and standard deviation (Bottom) of the joint-wise deviations of SMART and l2.}
  \label{fig:userStudy}
\end{figure*}

\subsubsection{Results.}

The success rate of {\bf Deceitfulness} is 93.32\% overall, which means that most of the time SMART does not visually change the semantics of the motions. When looking into the success rate on different datasets, SMART achieves 86.77\% on HDM05 and 96.38\% on MHAD. This also shows that most of the time people can tell different actions by observing skeletal motions, even for similar actions. Next, Figure \ref{fig:userStudy} Left shows the results of {\bf Naturalness}. Users' preferences over different losses are SMART $>$ l2-acc $>$ l2 $>$ l2-bone. SMART leads to the most natural results as expected.



Finally, we conduct the {\bf Indistinguishability} test. The final results are 81.9\% on average, 80.83\% on HDM05 and 83.97\% on MHAD. Note that this is a side-by-side comparison and thus is very harsh. The users are asked to find any visual differences. To avoid situations where motions are too fast to spot any differences (e.g. kicking and jumping motions), we also play the motions three times more slowly than the original.  Even under such harsh tests, humans still cannot spot any difference most of the time.

\subsection{Classifier Robustness under SMART Attack}

\begin{figure}[tb]
  \centering
  \includegraphics[width=0.45\textwidth]{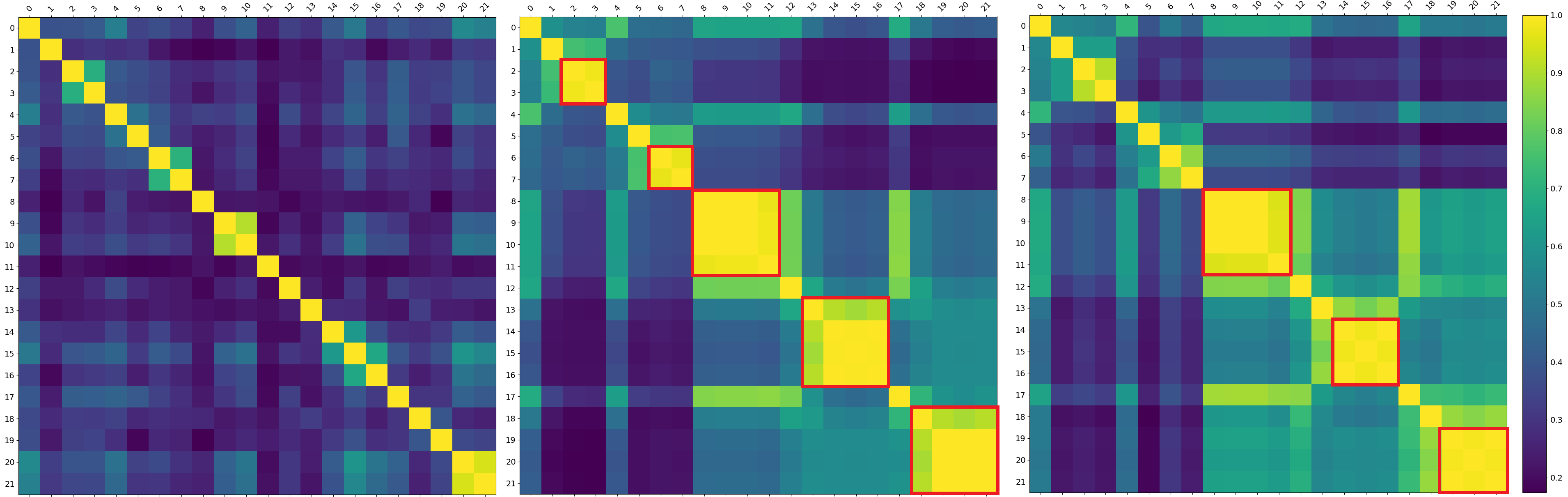}
  \caption{2S-AGCN on HDM05, displacement-displacement correlations (Left), displacement-speed correlations (Middle) and displacement-acceleration correlations (Right).}
  \label{fig:pertCorr}
\end{figure}

After rigorously confirming the effectiveness of SMART across datasets and models, we analyze the results to investigate the vulnerability of the target models. We start by looking at which joint or joint groups are attacked the most. Initially, if some joints tend to be attacked together, the correlations between the joint perturbations should be high. So we compute the Pearson correlations of joint perturbations, shown in Figure \ref{fig:pertCorr} Left. Although some local high correlations can be found (e.g. between joint 2 and 3, 6 and 7, 9 and 10, 20 and 21), they are not universal. Please see other results in the supplementary material. Next, we assume that the attack behavior might be class-dependent, i.e. depending on actions. However, after computing the joint perturbation correlations based on actions, no consistent and obvious patterns is found either.


Finally, we find that the displacement-speed and displacement-acceleration correlations reveal a consistent description of the vulnerability, shown in Figure \ref{fig:pertCorr} Mid and Right. The correlations are computed between the joint displacements and the original velocities and accelerations, respectively. These two correlations reveal the joint vulnerability: the higher the speed/acceleration is, the more the joint is attacked (shown by the high values along the main diagonal). In addition, they also reveal some consistent across-joint correlations (as shown by red boxes). Note that the joints in a red box belong to one part of the body (four limbs and one trunk). These joints normally have high within-group correlations in motions. Coordinated attacks on them easily fool the action recognizers.

The analysis suggests that joints with high velocity and acceleration are important features in the target models because these joints are attacked the most. This is especially so for joint groups with high within-group correlations. Most of the tested models are very sensitive to perturbations to these features, raising a big concern. Meanwhile, the analysis also suggests that reducing the sensitivity of a classifier over these features will increase its resistance to adversarial attack. To this end, one possible solution is to induce noises around the perturbation gradient during training, instead of purely white noises used by many methods. Another possibility is to introduce semantic descriptors (e.g. featuring a waving motion as one hand moving side-to-side above the head) which are not sensitive to small changes in these raw features.

\textit{Dynamics in Attack Imperceptibility}. To investigate the role of dynamics compared with joint-only perturbation, we conduct further analysis on SMART-vs-l2 where users prefer SMART to l2. We first compute their respective joint-wise deviations from the original motions, shown in Figure \ref{fig:userStudy} Right. In general, the perturbations of SMART are in general higher than l2 and have larger standard deviations. However, the users still choose SMART over l2. It indicates that with proper exploitation of dynamics, larger perturbations can generate even more desirable results. This is somewhat surprising and significantly different from the static data (e.g. images), where it is believed that the perturbation magnitude is tightly tied to imperceptibility~\cite{DBLP:journals/corr/abs-1907-10823}. This also suggests that classifiers could use perturbations on the dynamics to make the training more robust, which is complementary to the afore-mentioned suggestion of inducing noises around the perturbation gradient.

\subsection{Comparison}
To show that SMART is an effective tool for attack analysis, we compare SMART with IAA~\cite{8851936} and CIASA~\cite{Liu:2019:AAS}. As there are two competing factors (attack success vs imperceptibility), we fix one and compare the other. The success rate is largely governed by the clipping threshold of the perturbation magnitude in IAA and CIASA, and is hence easily tunable, while user studies on imperceptibility are expensive. We, therefore, tune IAA \& CIASA to achieve similar success rates, then conduct perceptual studies for comparison. Specifically, we conduct AB attack on HDM05 and the Indistinguishability test, as AB is also used in both papers. Each experiment includes 120 pairs of motions including motions evenly sampled from the original motions, SMART, IAA and CIASA results (30 motions each). In each trial, the left motion is the original motion while the right one is either the original motion, a SMART sample, an IAA sample or a CIASA sample. Results are shown in Table \ref{tab:comparison}. While the attack success rates of the three methods are similar, SMART, in general, generates more indistinguishable adversarial samples than IAA and CIASA do. We notice that most failures of IAA and CIASA are caused by broken motion dynamics and are therefore easily perceivable. This is understandable because IAA does not consider dynamics and thus generates jittering motions; CIASA uses GANs to govern the motion quality, which can only generate plausible motions, but not imperceptible samples. Details can be found in the supplementary materials.

\begin{table}[tb]
	\centering
    \begin{tabular}{c|ccc}		
		\hline
		Model/Method & SMART & IAA  & CIASA              \\ \hline
		HRNN                     & 100\%  & 98.12\% & 98.75\%        \\
		STGCN                     & 99.57\%   & 99.57\% & 99.56\%      \\
		2S-AGCN                     & 99.18\%  & 98.77\% & 98.98\%     \\
		\hline
        \hline
		HRNN                     & \textbf{42.22}\%  & 36.67\% & 32.22\%        \\
		STGCN                     & \textbf{90.00\%}   & 87.5\% & \textbf{90.00\%}      \\
		2S-AGCN                     & \textbf{80.83\%}  & 35.33\% & 49.33\%     \\
		\hline
	\end{tabular}

	\caption{Success rate in attack (Upper) and Indistinguishability (Lower). The attack success rate is the best results for SMART, IAA and CIASA.}
	\label{tab:comparison}
\end{table}


\section{Discussion}
Imperceptibility is vital in adversarial attack. When it comes to skeletal motions, perceptual studies are essential because there is no widely accepted metrics that fully reflect perceived realism/naturalness/quality. In addition, it helps us to uncover a unique feature of attacking skeletal motions. Losses solely based on perturbation magnitude are often overly conservative because they are mainly designed for attacking static data and unable to fully utilize the dynamics. Next, forming the joint deviation as a hard constraint~\cite{Liu:2019:AAS} via clipping is not the best strategy. The threshold needs to be manually tuned and it varies based on data. Besides, our perceptual study shows that larger perturbations can be used if the dynamics are exploited properly.

SMART is a straightforward but surprisingly effective attack method across datasets, models, attack strategies, and harsh perceptual studies. The simplicity of SMART raises an alarming concern for current action recognition research as it does not require complex computation to attack the state-of-the-art models. Through analysing SMART's behavior, we identified one key cause of their vulnerability: the over-sensitivity to joints with high velocity and acceleration, which we hope will help the future research to improve the recognition robustness.

\section{Conclusion and Future Work}
We demonstrated the vulnerability of several state-of-the-art action recognizers under adversarial attack. To this end, we proposed a new method, SMART, to attack action recognizers based on 3D skeletal motions. Through comprehensive qualitative and quantitative evaluations, we showed that SMART is \textit{general} across multiple state-of-the-art models on various benchmark datasets. Moreover, SMART is \textit{versatile} since it can deliver both white-box and black-box attacks with multiple attacking strategies. Finally, SMART is \textit{deceitful} as verified in extensive perceptual studies. Based on SMART, we revealed possible causes of the vulnerability of several state-of-the-art models. In the future, we would like to theoretically investigate why the transferability varies between different models under the black-box attack. We will also investigate how to systematically resist adversarial attack.

\noindent
\textbf{Acknowledgements:} We thank Qun-Ce Xu and Kai-Wen Hsiao for their help on the perceptual study. This   project   has   received   funding   from   the   European
Union’s Horizon 2020 research and innovation programme
under grant agreement No 899739 CrowdDNA, EPSRC (EP/R031193/1), NSF China (No. 61772462, No. U1736217), RCUK grant CAMERA (EP/M023281/1, EP/T014865/1) and the 100 Talents Program of Zhejiang University.

{\small
\bibliographystyle{ieee_fullname}
\bibliography{egbib}
}

\end{document}